%% file: main.tex
\documentclass[10pt,twocolumn,letterpaper]{article}

\usepackage{cvpr}      % To produce the REVIEW version

\usepackage{graphicx}
\usepackage{amsmath}
\usepackage{amssymb}
\usepackage{booktabs}
\usepackage{mathtools}
\usepackage{multirow}
\usepackage{float}
\usepackage{placeins}
\usepackage[accsupp]{axessibility}
\usepackage{nopageno}

\usepackage[pagebackref,breaklinks,colorlinks]{hyperref}

\usepackage[capitalize]{cleveref}
\crefname{section}{Sec.}{Secs.}
\Crefname{section}{Section}{Sections}
\Crefname{table}{Table}{Tables}
\crefname{table}{Tab.}{Tabs.}

\makeatletter\renewcommand\paragraph{\@startsection{paragraph}{4}{\z@}
  {.5em \@plus1ex \@minus.2ex}{-.5em}{\normalfont\normalsize\bfseries}}\makeatother

\usepackage{xcolor}

\def\cvprPaperID{5942} % *** Enter the CVPR Paper ID here
\def\confName{CVPR}
\def\confYear{2022}

\begin{document}

\title{SwapMix: Diagnosing and Regularizing \\ the Over-Reliance on Visual Context in Visual Question Answering}

% \title{VQA models Overrely on Irrelevant Context; \\Swapping Context Features Improves Accuracy and Robustness}

\author{%
Vipul Gupta\textsuperscript{$1$} \enspace Zhuowan Li\textsuperscript{$2$}  \enspace Adam Kortylewski\textsuperscript{$2$} \enspace Chenyu Zhang\textsuperscript{$2$} \enspace Yingwei Li\textsuperscript{$2$} \enspace Alan Yuille\textsuperscript{$2$} \\
% \vspace{.3em}
{\textsuperscript{$1$} Pennsylvania State University    } \qquad \quad
{\textsuperscript{$2$} Johns Hopkins University    } \qquad \quad \\
% \vspace{-.25em} 
{\tt \small vkg5164@psu.edu \enspace \{zli110, akortyl1, czhan129, yingwei.li, ayuille1\}@jhu.edu}
}
\maketitle

\input{content/00_abstract}
\input{content/01_introduction}

\input{content/02_related_work}
\input{content/03_method}

\input{content/04_experiments}
\input{content/05_conclusion}
\input{content/06_acknowledgments}
\clearpage
\newpage
{\small
\bibliographystyle{ieee_fullname}
\bibliography{references}
}

\end{document}

%% file: content/00_abstract.tex
\begin{abstract}
While Visual Question Answering (VQA) has progressed rapidly, previous works raise concerns about robustness of current VQA models. %further analysis shows that these models are quite vulnerable to adversarial attacks.
In this work, we study the robustness of VQA models from a novel perspective: visual context.
%present a new way to benchmark VQA models and
We suggest that the models over-rely on the visual context, i.e., irrelevant objects in the image, to make predictions.
% Your details part
To diagnose the models' reliance on visual context and measure their robustness, we propose a simple yet effective perturbation technique, SwapMix. 
SwapMix perturbs the visual context by swapping features of irrelevant context objects with features from other objects in the dataset.
% Experiments
Using SwapMix we are able to change answers to more than 45\% of the questions for a representative VQA model.
Additionally, we train the models with perfect sight and find that the context over-reliance highly depends on the quality of visual representations. 
% This is because improved quality of visual inputs helps the model to focus on relevant objects.  
% Introduce swapmix 
In addition to diagnosing, SwapMix can also be applied as a data augmentation strategy during training in order to regularize the context over-reliance. By swapping the context object features, the model reliance on context can be suppressed effectively. 
Two representative VQA models are studied using SwapMix: a co-attention model MCAN and a large-scale pretrained model LXMERT. Our experiments on the popular GQA dataset show the effectiveness of SwapMix for both diagnosing model robustness, and regularizing the over-reliance on visual context. The code for our method is available at \href{https://github.com/vipulgupta1011/swapmix}{\textcolor{magenta}{https://github.com/vipulgupta1011/swapmix}}
% SwapMix reduces over-reliance on context and forces the model to focus on relevant objects. 
% SwapMix increases the effective accuracy of the VQA models by atleast 1.3\%. 
%Scene graph model accuracy and robustness
 
% We also train both VQA models using scene graphs and show that models trained on scene graphs have higher accuracy and are far more robust to perturbations than FasterRCNN models. 
\end{abstract}

%% file: content/01_introduction.tex
\section{Introduction}

% \zhuowan{Para1: Overview of VQA (attention, contextual information).}
Visual Question Answering (VQA) is a challenging task that requires a model to answer open-ended questions based on images. In recent years, VQA performance is greatly boosted by different techniques including intra- and inter-modality attentions \cite{yu2019deep, Cadne2019MURELMR}, large scale multi-modal pretraining \cite{tan2019lxmert, Li2019VisualBERTAS, li2020oscar}, etc. However, previous works study the robustness of VQA models and show that the models may exploit language prior \cite{VQA-CP_v2, Niu2021CounterfactualVA, Niu_2021_CVPR}, statistical bias \cite{agrawal2016analyzing, goyal2017making} or dataset shortcuts \cite{GQA-OOD, Kervadec2021HowTA} to answer questions.

%co-attention models \cite{yu2019deep, yu2017multi, Zhang2021MultimodalFC}, multimodal attention \cite{Cadne2019MURELMR, fukui2016multimodal}, transformer models \cite{Su2020VL-BERT:, Li2019VisualBERTAS, yu2018beyond}, large-scale pretrained models \cite{tan2019lxmert, Gao2019DynamicFW}, graph attention network \cite{Li2019RelationAwareGA}. All these models try to use the contextual information present in the image to answer questions.

% \zhuowan{Para2: Robustness of VQA, cite several previous works, from the perspective of language prior, statistical bias, common-sense, etc. We approach the problem from a different perspective: context.}

\begin{figure}[t!]
\begin{center}
%\fbox{\rule{0pt}{2in} \rule{.9\linewidth}{0pt}}
\includegraphics[width=1.0\linewidth]{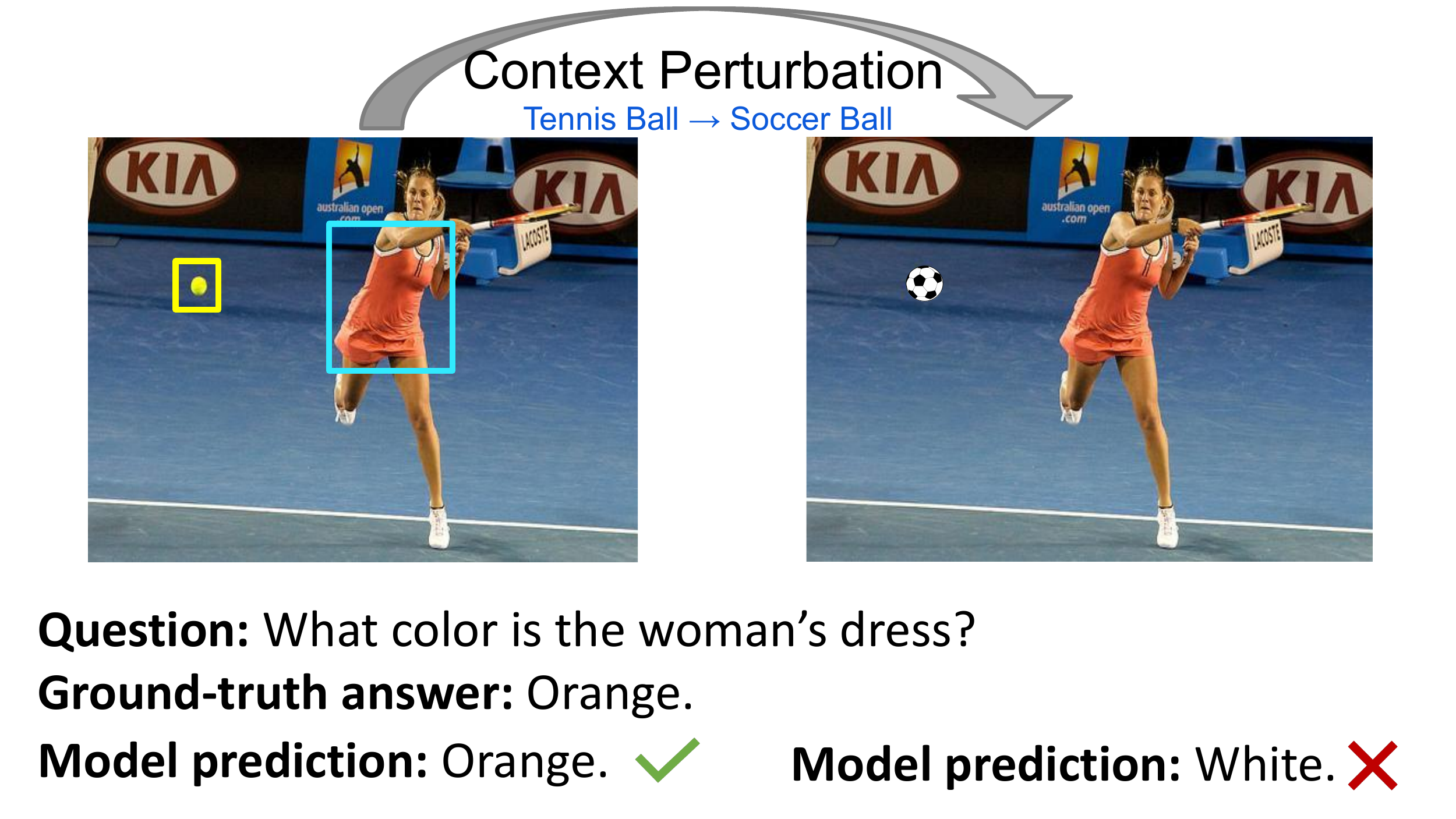}
\end{center}
\vspace{-1.5em}
   \caption{VQA models over-rely on visual context. By swapping features of irrelevant context objects, we can perturb the model prediction. Here the tennis ball (in \textcolor{yellow}{yellow} box) is an irrelevant context object for the question. Changing feature of the tennis ball to feature of soccer ball results in change in model prediction.}
\vspace{-1.0em}
\label{fig:intro}
\end{figure}

While previous works studied VQA robustness from the perspective of language context, in this work, we study the robustness of VQA models from a different view: visual context. The visual context refers to the background in the image or the irrelevant objects that are not needed during the reasoning process to answer the question. For example, in Figure \ref{fig:intro}, the tennis ball is irrelevant for the question "What color is the women's dress", so we say it is a context object. Ideally, a model with real perception and reasoning ability should be robust to the irrelevant context. However, in our work, we find that VQA models are vulnerable to context changes, which suggests the models' over-reliance on the irrelevant context in the image.

% \zhuowan{Para3: Briefly describe: we propose a simple perturbation strategy to randomize context that surprisingly attacks XXX\% of questions, which shows the over-reliance of context.}

To study the role of visual context, we propose a simple perturbation strategy named \textit{SwapMix}, which perturbs the visual context by swapping features of context object with features from another object in the dataset. We first identify the visual features corresponding to irrelevant objects in the image, then randomly swap them with feature vectors of another similar object from the dataset. For example, in Figure \ref{fig:intro}, the tennis ball is a context object for the given question, so we swap tennis ball feature vector with a feature vector of soccer ball. The swapping confuses the model to mis-recognize the color of the dress. In the swapping process, we carefully control the swapped objects to ensure that the new object is compatible to the scene (\eg, we don't want to change the ball into a car). 

Surprisingly, by perturbing the irrelevant context, more than 45\% of the correct answers get changed. This reveals that VQA models highly rely on the context in the image, thus are vulnerable to context perturbations. The model may utilize shortcut correlations in the visual context to make predictions. We diagnose two representative VQA models: MCAN \cite{yu2019deep} as representative for attention-based models, and LXMERT \cite{tan2019lxmert} as representative for large-scale pretrained models. Our experiments show that LXMERT is much more robust to context perturbations, which indicates that large-scale pretraining may increase model robustness.

We further find that the context over-reliance highly depends on the quality of visual representations: a perfect sighted model relies much less on context. We achieve this by replacing the visual representations \footnote{Majority of VQA models use object features extracted by pretrained object detectors as visual representation.} with the ground-truth object and attribute encoding, which can be viewed as gold visual representation that provides the model the perfect sight. By studying this perfectly sighted model, we can exclude the influence of imperfect visual perception, thus purely focus on the reliance on relevant objects in the reasoning process. Our results shows that by providing VQA models with the perfect visual encoding, the answer changes are greatly reduced from 45.0\% to 16.4\% (for MCAN model). This suggests that models trained with perfect visual representations are more robust and that the context over-reliance largely comes from the imperfection of visual perception features. 

% \zhuowan{Para4: Motivated by para3, we propose a strategy to overcome the context over-reliance. Summarize SwapMix and its experiment results.}

In addition to diagnosing context over-reliance, SwapMix can also be used as a data augmentation technique during training. In training, we randomly swap a part of the context features with other object features from the dataset. This forces the model to focus more on relevant objects in the image and less on irrelevant context. Our empirical results show that by applying \textit{SwapMix} in training, the model robustness improves by more than 40\% and effective accuracy improves by more than 5\% on GQA dataset \cite{GQA}.

% We use GQA dataset \cite{GQA} for our experiments as it provides detailed annotations of images along with reasoning steps for the question. Most of the VQA models uses FasterRCNN features \cite{Ren2015FasterRT} of images as visual input to the VQA model. Some of the recent works have used scene graphs for VQA tasks \cite{Ghosh2019GeneratingNL, Lee2019VisualQA}. We do our analysis on both faster rcnn features and scene graph embeddings to generate perturbations and SwapMix training. We found that models trained with scene graphs are more robust and have higher performance.

% \zhuowan{Para5: Summary of contributions}

Our main contributions in this paper are three-fold. 
First, we are the first to study VQA robustness from the perspective of visual context. With our simple context perturbation strategy named SwapMix, we benchmark robustness of two representative VQA models and find their over-reliance on visual context.
Second, we find that a perfect sighted model relies much less on visual context. We provide models with perfect visual encodings and observe the improvement in model robustness. Third, we define 2 metrics, context reliance and effective accuracy and shows improvement by using SwapMix as data augmentation technique.

%% file: content/02_related_work.tex
\section{Related Works}

\paragraph{Visual Question Answering.}
% We will first do a Visual Question Answering (VQA) method overview.
% VQA\cite{antol2015vqa} is a challenging task that .....
The most common approach for VQA is to first extract visual features using convolution neural networks and question features using LSTM\cite{antol2015vqa}, then fuse them together to make answer predictions \cite{zhou2015simple}. Multiple works have shown the effectiveness of attention in VQA \cite{chen2015abc, yang2016stacked, fukui2016multimodal, kim2016hadamard, yu2017multi, yu2018beyond, ben2017mutan}. BAN\cite{kim2018bilinear} proposes bilinear attention that utilizes vision and language information. MCAN \cite{yu2019deep} is a co-attention model which uses self-attention and guided-attention units to model the intra-modal and inter-modal interactions between visual and question input. OSCAR \cite{li2020oscar} uses object tags in images as anchors to improve alignment between modalities. LXMERT \cite{tan2019lxmert} is a large-scale Transformer \cite{vaswani2017attention} model that consists of three encoders: an object relationship encoder, a language encoder, and a cross-modality encoder. In concurrent work, \cite{cascante2022simvqa} proposes feature swapping for domain adaptation from synthetic to real data.

% NSM\cite{hudson2019learning} transformed both visual and linguistics modalities into semantic concept-based representations. 
% backbone models\cite{simonyan2014very, szegedy2015going, he2016deep} 

% VQAv2\cite{goyal2017making}
% VQA-CP\_v2\cite{VQA-CP_v2}
% CLEVR\cite{CLEVR}
% VQA Genome\cite{VQA_Genome}
% GQA\cite{GQA}

\paragraph{Biases and Robustness in VQA.}
Despite the prosperity in the development of VQA, multiple previous works show bias in VQA models. \cite{agrawal2016analyzing} points out the generalization incapacity of VQA models. \cite{hendricks2018women} shows bias reliance of VQA models. \cite{manjunatha2019explicit} discover and enumerate explicitly biases learned by the model. Many work show that the models exploit language prior \cite{VQA-CP_v2, Niu2021CounterfactualVA, Niu_2021_CVPR}, statistical bias \cite{agrawal2016analyzing, goyal2017making} or dataset shortcuts \cite{GQA-OOD, Kervadec2021HowTA} to answer questions.
% Ways to decrease biases:
There are many approaches to mitigate the bias in models. \cite{VQA-CP_v2} introduces a method that reorganizes the VQA v2 dataset. Some works use question-only model: \cite{ramakrishnan2018overcoming} introduces training as an adversarial game between the base model and a question-only adversary, while \cite{RUBi} adds a question-only branch to do joint training with the base model, and omits it at test time. CSS \cite{CSS} generates counterfactual samples during training, which improves the visual-explainable ability. \cite{wu2019self,selvaraju2019taking} leverage the important visual information by humans to focus on selected regions during training. \cite{clark2019don} designed a two-stage model, the first stage trains only on biases, and the second stage focuses on the other patterns of the dataset. 
% Ways to measure biases:
In addition to decreasing modal biases, there are lot of work on measuring biases more accurately and efficiently.
MUTANT \cite{gokhale2020mutant} and GQA-OOD \cite{GQA-OOD} use out-of-distribution (OOD) generalization. Early work like \cite{malinowski2014multi} provided a soft measure score based on a lexical set. 
% NSM \cite{hudson2019learning} introduced ways to split the training set to measure generalization of content and structure.
\cite{bahdanau2019closure} measures the performance of the models based on both the baseline questions and the CLOSURE test, indicating that the gap between these two measurements is the behavior of generalization. \cite{Dancette2021BeyondQB} measures bias in VQA by finding counter-examples from validation set with their proposed rules and use the mined counter-examples to evaluate model.
Different from the above previous works, our work is the first to study the reliance on visual context of VQA models by generating new examples.

\paragraph{Context in computer vision.}
Contextual information is important for computer vision. For object recognition, early work by \cite{torralba2003context} introduced a context-based model using place categorization to simplify object recognition, \cite{oliva2007role} studied how context influences object recognition, and recent work by \cite{zhong2020cascade} modified a global context model to enhance performance. 
Moreover, \cite{wang2020robust} demonstrate that object detection models rely too much on contextual information when objects are occluded, and resolve this using a compositional generative model \cite{kortylewski2020compositional,kortylewski2021compositional} that separates the object and context in the representation.
For scene graphs, \cite{ren2020scene,wang2020sketching} introduce a hierarchical context model to generate a scene graph, and \cite{lin2020gps} augment the node features of scene graphs with contextual information. For segmentation, \cite{he2019adaptive} presents multi-scale contextual representation with context modules, which leverage the global image representations to estimate local affinity of sub-regions, and \cite{lin2018scn} introduces a switchable context network to improve the performance of semantic segmentation of RGB-D images. In the field of VQA, \cite{sharma2021visual} add a visual context based attention that takes into account the previously attended visual content. 
% In this work, we measure robustness of VQA model using context perturbation.

% These works unveil the importance of context in VQA models, which has great influences on biases, robustness and accuracy. 

%% file: content/03_method.tex
\section{Method}

\begin{figure*}[t!]
\vspace{-0.5em}
\begin{center}
%\fbox{\rule{0pt}{2in} \rule{.9\linewidth}{0pt}}
\includegraphics[width=0.95\linewidth]{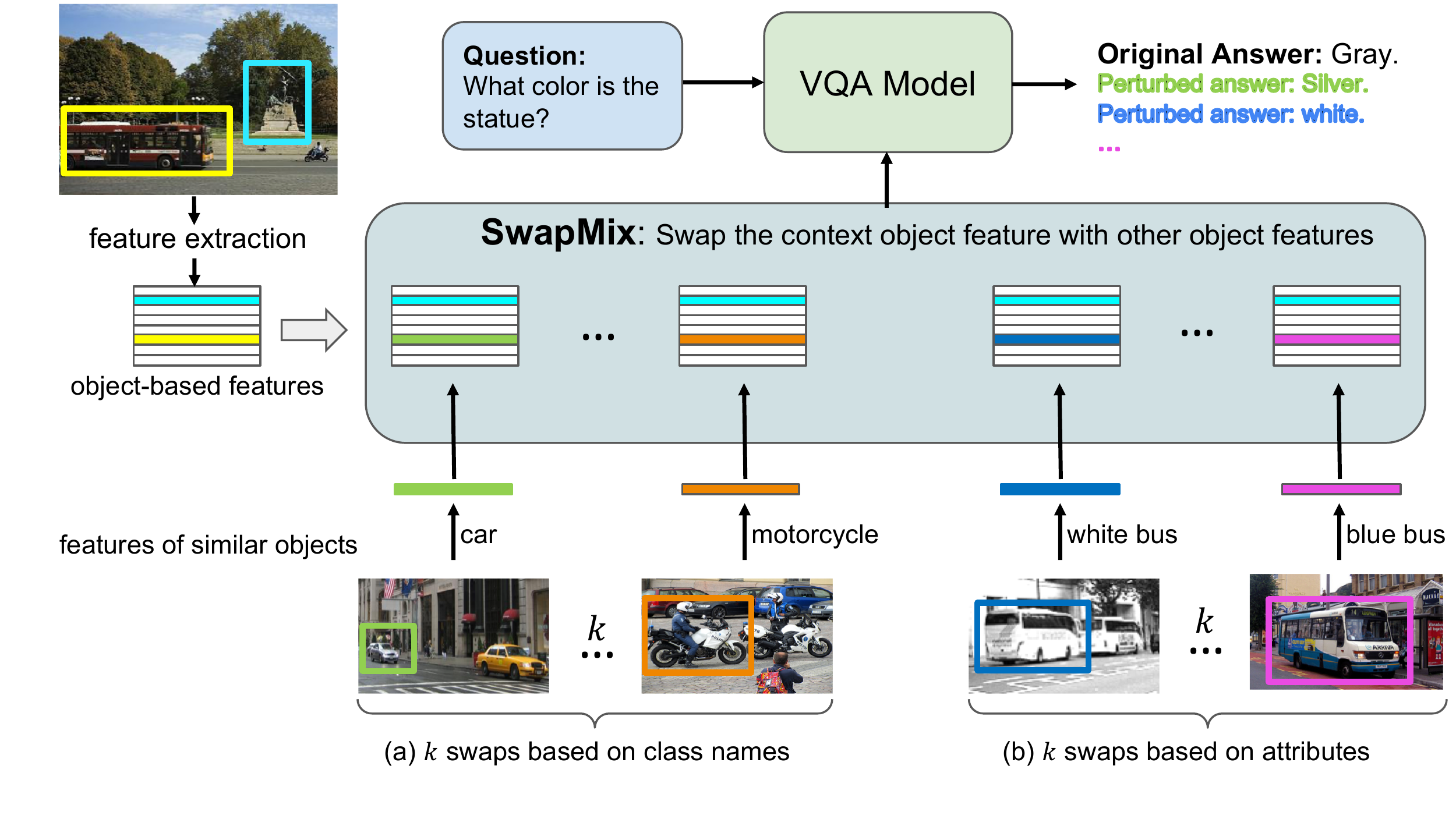}
\end{center}
\vspace{-3em}
   \caption{Overview of our method. Given an image and a question, we first find context object (\eg red bus in the yellow box) using the reasoning steps of the question. Then we swap the context object feature with other similar object features in the dataset. We perform $k$ swaps based on (a) object class names and (b) object attributes each. The model's reliance on context can be evaluated with the percentage of answer changes when context gets perturbed.}
\vspace{-1em}
\label{fig:method}
\end{figure*}

% \vipul{Para1: We have used GQA dataset. Explain why we used GQA.}
% \zhuowan{Don't say anything in the method part, only describe GQA in experiment section.}

% \vipul{Para2: Overview of feature swapping technique - swapping irrelevant context in the image with features from the dataset.}
% \zhuowan{In this para, also highlight the motivation for doing this}

% \vipul{Para3: Describe 2 ways of perturbations : object and attribute}
% \zhuowan{Say the object and attributes only in one sentence in Para2, no need to highlight this using on para here. In para3, we can talk about: (level 1) we first do perturbation to diagnose the model, then (level 2), based on the perturbation findings we propose SwapMix training stragegy to improve the model.  }

VQA models are not robust to minor perturbations. In this section, we provide a simple perturbation technique that measures the reliance of VQA models on visual context, i.e. irrelevant objects image. We swap features corresponding to irrelevant objects in the image with other objects from the dataset. In an ideal scenario, changing the context objects in the image should not affect the model's prediction, while in our experiments, we found that VQA models rely heavily on the context and are not robust to small perturbations.

We name our method, SwapMix, which performs perturbations on visual context to diagnose the robustness of the model. SwapMix can also be used as a data augmentation technique during training to improve the robustness and effective accuracy of the model. We first define what visual context is, then introduce VQA models with perfect sight which leads to interesting diagnosing findings, next describe how we perform SwapMix, and finally talk about how to apply SwapMix as a training strategy.
% \subsection{FasterRCNN}
% \vipul{Para4: Explain how features are selected based on IoU ratio and how swapped features are selected based on topk matches and random selection}

% \vipul{Para5: Introduce hyperparameter N, no of perturbations}

% \subsection{Scene Graphs}
% \vipul{Para6: Describe how we trained the model. Using glove embeddings and bounding boxes in place of Faster RCNN features}

% \vipul{Para7: Tell how feature swapping works and hyperparameter N, same as faster rcnn}

% \vipul{Para8: Some mathematical equations to show the problem formulation.}

\subsection{Definition of Visual Context} \label{sec:defi_context}

Here we clarify the definition of visual context and provide formulation for the problem. 
% VQA models takes an image and a question as inputs and output answer prediction:
$$ f = Model(V, Q)$$

Here, $V$ represents the visual representation and $Q$ represents the question input. A widely-used visual representation is the object-based features \cite{anderson2018bottom} extracted by pretrained object detector Faster RCNN \cite{Ren2015FasterRT}. In this case, $V \in \mathbb{R}^{n \times d}$ is a set of object features, where $n$ is the number of objects in the image and $d$ is the dimension of the feature vector for each object.

Among the $n$ objects in the image, there are some irrelevant objects that are not needed in the reasoning process of question answering. For a fully robust model, changing the context, $C$ should not change model's prediction as shown in Figure \ref{fig:intro}.
% For example, in Figure \ref{fig:intro}, for question “What color is the women's dress”, tennis ball is an irrelevant object while dress is a relevant object. 
We refer to those irrelevant objects as visual context and denote visual context by $C$. $C \in \mathbb{R}^{m \times d}$ is a subset of $V$. It contains feature vectors corresponding to $m$ irrelevant objects. Each row of the context $C$, denoted as $\mathbf{c}_{i}$, is a feature vector corresponding to an irrelevant object. 
% \zhuowan{Para1: briefly describe what context is, and math denotation of relevant features and irrelevant ones (context)}

The context objects are identified using the question reasoning steps. For example, in order to answer the question ``What color is the statue in front of the trees", we need to first find the tree, then find the statue in front of the tree and finally query its color. The GQA dataset \cite{GQA} provides the ground-truth reasoning steps for each question, as well as the selected objects after each step. We use those reasoning steps to filter out all the relevant and irrelevant objects for the question.
%We select the objects used at each step of reasoning and add them to relevant objects. For example for the question "Is there a red apple on the table?", reasoning steps are \{"operation": "select", "argument": "table (279472)"\}, \{"operation": "relate","argument": "on, subject, apple (271881)"\}, \{"operation": "filter", "argument": "red" \},\{ "operation": "exist","argument": "?"\}.
% \zhuowan{Make this paragraph more specific: what is reasoning steps and how it is used to filter out the context.}. 
% For this question, we pick table (279472) from step 1 and apple (271881) from  step2 as relevant objects. All the other objects in the image belong to the context, $C$. 
Then Intersection-over-Union (IoU) ratio is used to match the predicted objects with the ground-truth ones.
% . \zhuowan{Provide the IOU threshold in experiment details. BTW, is it threshold or top-1?} \vipul{it is top-1}

% \zhuowan{Para2: (on ground-truth/scene graph level) how do you discover target/context? e.g get the reasoning steps of the questions, filter out the used/unused objects for each step, etc}

% \zhuowan{Para3: after you get the GT contexts as in Para2, how do you get that for FasterRCNN features? e.g IOU computing etc}

\subsection{VQA Model with Perfect Sight} \label{sec:perfect_sight}

We conjecture that the model robustness is related to the quality of visual perception. The majority of the current VQA models use the object features described above as visual input to the model. The features are extracted by a pretrained off-the-shelf object detector which is not updated in VQA training. These pre-extracted features may contain a large amount of noise and miss out on important information that is required to answer the question. In this case, the model may be forced to learn unreasonable data correlations from irrelevant context to predict the answers correctly, which reduces the robustness of the model.

Therefore, to study the influence of visual perception imperfection, we train a model with perfect sight and compare its behavior with model trained with commonly used detected features. 
% Some of the recent works \cite{Ghosh2019GeneratingNL, Lee2019VisualQA} have used scene graph embedding as visual input in place of Faster RCNN features. In scene graphs representation, each image comes with a graph of objects and relations. Scene graph representations are converted to features using word embeddings. Advantage of using scene graphs is that scene graphs are easy to perturb and their embeddings contains less noise. On the other hand, datasets with scene graph representations are very limited. In this work, we perform perturbations on both Faster RCNN features and scene graphs to measure effect of context on predictions by VQA models.
% \vipul{Make this para more clear}
The models with perfect sight are trained using the scene graph annotations in GQA dataset. We replace the object features with the encoding of ground-truth object annotations. More specifically, for each object, we encode its annotated class label and attributes into one-hot encodings, which are then encoded using GloVe \cite{pennington2014glove} embeddings and finally converted to inner dimension $d$ with a FC layer. The object bounding box coordinates are also converted to same dimension using a FC layer. The final representation of an object $i$ is the average of three parts: $\mathbf{c}_{i} = Avg(\mathbf{o}_{i}, \mathbf{a}_{i}, \mathbf{b}_{i})$. $\mathbf{o}_{i}, \mathbf{a}_{i}, \mathbf{b}_{i} \in \mathbb{R}^{1 \times d}$ are encodings for object class label, attributes and bounding box coordinates respectively. 
% $O_{G}  \in \mathbb{R}^{1 \times 300}$. We get feature vector for attributes by taking average of Glove embeddings for $l$ attributes, $A_{G}  \in \mathbb{R}^{1 \times 300}$. Bounding box coordinates for the object are also transformed to same dimensionality using a fully connected layer. $B_{G}  \in \mathbb{R}^{1 \times 300}$. Then we take average of \(O_{G}, A_{G}, B_{G}\), which serves as the feature vector for the object. Feature vector for each object in scene graphs, $C_{object}  \in \mathbb{R}^{1 \times 300}$ 
% Length of one-hot encoding for class label is the number of classes in the entire dataset. One-hot encoding for class label will have 1 corresponding to its label and 0 for all other labels. Similarly, for one-hot encodings of attributes we have 1 corresponding to each attribute in the object and 0 for the rest. 

\subsection{SwapMix} \label{sec:swapmix}
% \zhuowan{First paragraph to have a high-level overview: introduce what it is; say it can be applied in both diagnosing and training; then introduce the structure of this section}

% \zhuowan{Everything before the More Details paragraph should be based on image features. Put everything specific to scene-graph into the More Details paragraph.}

% VQA models exploit the visual and language biases for improving accuracy \cite{hendricks2018women, manjunatha2019explicit}. They use context of the image for answering the questions. 
Now we introduce our proposed context perturbation strategy: SwapMix. The overall idea is shown in Figure \ref{fig:method}. First, we describe the broad idea about the method and then we go into details on how we select candidates for context swapping, how we perform context swapping in terms of object class labels and attributes, and finally, how we apply SwapMix as a training strategy to improve robustness of the models. 

% which measures the effect of context on VQA models. SwapMix can be used both for diagnosing a trained model and regularizing the over-reliance on the context by using it as data augmentation during training. In this section, f

% Every row of $C$ is a feature vector of an irrelevant object which corresponds to a class label and its attributes. For example, in Figure \ref{fig:intro}, an irrelevant object can be defined as class\_label=\{tennis ball\}, attributes=\{green, round\} \zhuowan{Use this denotation: $\mathbf{c}_{i}(tennis\_ball; green, round)$. Define this denotation in the context definition subsection, $\mathbf{c}_{i}(o;a_{1}, a_{2}...)$}. 
After discovering the context objects as described in Section \ref{sec:defi_context}, we swap their features with other object features from the dataset. For each context object, we perform two types of context swapping based on (a) class label and (b) attributes. In (a) we swap the object feature with the feature of an object from a different class. For example, we change a \textit{bus} into a \textit{car}. In (b) we swap the object feature with feature of an object from same class but with different attributes. For example, we change a \textit{red bus} with \textit{yellow bus}. For both (a) and (b), we perform $k$ feature swaps per context object, therefore altogether we have $2k$ swaps for each context object. We perform context swapping iteratively for each irrelevant object in the image and measure the percentage of answer changes.

% We conduct perturbation on feature level instead of directly modifying the image 

We control the swapping process to make sure that the new object is compatible with the image. For example, we may want to change a \textit{bus} into a \textit{car}, but we don't want to change it into a \textit{computer} because a \textit{computer} parking on the roadside is unnatural. swapped feature always corresponding to an object from the dataset. The swapped feature resembles a real object and thus this perturbation is equivalent to replacing the irrelevant object with another object in the image.
Next, we will provide more details on how we choose candidate features to swap with.

To better describe our feature swapping strategy, we denote each context object feature (each row of context $C$) as $\mathbf{c}_{i}(o;a_{1}, a_{2}, \ldots)$. Here $o$ is the class label for the object $i$, and $a_{1}, a_{2}, \ldots$ are its attributes. Each object belongs to a unique object class while it can have an arbitrary number of attributes. For example in Figure \ref{fig:method}, feature corresponding to the large red bus can be written as $\mathbf{c}(bus;\: red,\, large)$.
% Each object has its class label and attributes. Each row of context $C$, $\mathbf{c}_{i}(o;a_{1}, a_{2}...)$, is a feature vector corresponding to a context object $o$.   It is represented

% As we do not change the features of relevant objects for the question, context swapping changes the features of the image but does not affect the answer. 
% We use a hyperparameter $k$, which corresponds to number of changes/swaps per object belonging to context.

% We make sure that the swapped feature corresponds to a real object. We will provide more details on this in the following paragraphs.

% \zhuowan{Then in this paragraph, describe how you get candidates for object name and attribute swapping.}
% \paragraph{Candidates for context swapping.}
% We filter m irrelevant objects corresponding to question from the reasoning steps as explain in the above section. For each irrelevant object, we perform two types of context swapping based on (a) class label and (b) attributes. 

% we make a list of similar attributes in the similar way using GloVe embeddings and cosine similarity. 

\paragraph{Swapping the context class.}
% \zhuowan{Here use a paragraph to make it clear that we are doing perturbations/swapping on two levels: (a) object names; (b) object attributes. Describe what it means by swapping object names or attributes here.}
% \zhuowan{Enrich this paragraph and the following one by merging in the paragraph Perturbation on image features/scene graphs. } 

We swap the object feature with the feature of an object from a different class. For example in Figure \ref{fig:method} we swap the \textit{bus} to \textit{car}, \textit{motorcycle}, etc. This type of feature swapping is similar to putting an object of a different class in place of the irrelevant context object in the image. The context swapping helps us understand the dependency of the VQA model on irrelevant objects in the visual input. 

To ensure that the swapped class is in the similar domain as object of interest, we only swap the object into similar classes. To achieve this, for each context object, we find the $k$ nearest classes to its class name. More specifically, we compute cosine similarities between GloVe embeddings of class names and pick the top $k$ class labels. Additionally, we set a threshold (0.5) to filter out the classes with small similarities \footnote{For cases where the number of matching classes with threshold 0.5 are less than $k$, we select random classes from the datasets to make number of swaps exactly $k$. We also tried not padding with random classes. The experimental results are similar for the two settings.}. In this way, we get the candidates for swapping each context object and ensure the selected matching classes are in the similar domain. For example, for the class \textit{car}, its top-5 similar classes are: \textit{truck, motorcycle, vehicle, taxi, bus}.
%, train car, suv, toy car, minivan, jeep, food truck, bus stop\}. 

For each context object $\mathbf{c}_{i}(o;a_{1}, a_{2}, \ldots)$ with class $o$, we select the $k$ nearest class labels to $o$ for swapping as described above. For each of the $k$ classes, we randomly pick one object from the dataset that belongs to that class. This results in $k$ candidate features for swapping: $\mathbf{\hat{c}}_{i}^{j}(o_{j};a_{1}^{j}, a_{2}^{j}, \ldots)$ where $j=\{1,2....k\}$. Then by swapping $\mathbf{c}_{i}$ to each of the $\mathbf{\hat{c}}_{i}^{j}$, we get $k$ perturbations for each irrelevant context object.
% \{$l_{1}, l_{2},...., l_{k}$\}

To perturb the perfectly sighted models trained with the encoding of ground truth object class and attributes, a straight-forward way is to simply modify the one-hot encoding for the class label. However, this might cause the object names to be incompatible with attributes, such as \textit{pink elephant} or \textit{green basketball}. Therefore, to ensure that the swapped context corresponds to a real object, we pick a random object of the swapped class from the dataset and use its attributes to generate one-hot encodings for the swapped object attributes. 

% \[ cosine(C_{i},Y) = \dfrac{C_{i} \; . \; Y}{|C_{i}| \; . \; |Y|}\]

% We select a threshold of 0.5 to ensure the top $k$ selected matching classes are in similar domain. For cases where the number of matching classes with threshold 0.5 are less than hyperparameter $k$, we select random classes from the datasets to make number of swaps exactly $k$. This checks the robustness of the model by replacing context with classes which are from a different domain. 
% We provide analysis with both including and not including random class selection.

\paragraph{Swapping the context attributes. }
To further study the context reliance, we change the object attributes while keeping the object class unchanged. The object feature is swapped with the feature of an object from the same class but with different attributes. For example in Figure \ref{fig:method}, the \textit{red bus} can be changed to \textit{orange bus}, \textit{yellow bus}, etc. Compared with object class swapping, attribute swapping can be viewed as a more controlled perturbation that helps reveal the models' reliance on the context in more detail.

For each context object $\mathbf{c}_{i}(o;a_{1}, a_{2}, \ldots)$, we swap it with $k$ objects of same class label but different attributes. To get the $k$ swapping candidates, we randomly select $k$ objects from the list of objects belonging to same class with different attributes from the dataset \footnote{If the list of objects belonging to same class is less than $k$, let's say $k'$, we perform only $k'$ feature swapping for that object.}. This results in $\mathbf{\hat{c}}_{i}^{j}(o;a_{1}^{j}, a_{2}^{j}, \ldots)$ where $j=\{k+1,k+2, \ldots, 2k\}$. The object $o$ remains the same across all context swaps.
% This is because in this category of swapping, we want to test robustness of the model to change in attributes only. 
% For example in Figure \ref{fig:intro}, we can replace $\mathbf{c_{i}}(tennis\, ball;\, green,\, round)$ to $\mathbf{c_{i}}(tennis\, ball;\, red,\, round)$. This type of feature swapping is equivalent to putting a different object of same class in place of the irrelevant object in the image. This helps us to measure the dependency of the VQA model on attributes belonging to irrelevant objects in the visual input.

To perturb the model with perfect sight, we just change the one-hot encodings for the attributes. We pick top $k$ attributes which are similar to attribute of interest using GloVe similarity. For example, the attribute \textit{black} can be swapped with: \textit{blue, green, red, purple, yellow}.% green, brown, dark blue, white, black and white\}.

\paragraph{Algorithm.}
% \zhuowan{I feel like this part is actually implementation details. Maybe we do not need to describe those details?} 
Let $\mathbf{c}_{j} \in \mathbb{R}^{1 \times d}$ be the context corresponding to ${j}^{th}$ irrelevant object. The aim is to swap $\mathbf{c}_{j}$ with context $\mathbf{c}_{p}(o^{p};a_{1}^{p}, a_{2}^{p}, ...)$ belonging to an object $p$ from the dataset. We define a matrix for swapping, $S \in \mathbb{R}^{m \times d}$, where each row of S is equal to $\mathbf{c}_{p}$. We perform feature swapping to convert $C$ to $C^{p}$ with the following operation : 

\[ C^{p} = C \odot P + S \odot P^{c} \]

$P \in \{0,1\}^{m \times d}$ is the perturbation matrix and \( \odot \) is Hadamard product \cite{million2007hadamard}, also known as element wise matrix multiplication. All the rows of P are 1's except ${j}^{th}$ row corresponding to ${c}_{j}$. $P^{c}$ is complementary matrix of P, $P^{c}$ = J -  P, where J is the matrix with each entry being 1. Thus, all entries of $P^{c}$ are 0's except for ${j}^{th}$ row. Effectively, we modify the context $C$ to $C^{p}$ by changing context of ${j}^{th}$ row, from $\mathbf{c}_{j}$ to $\mathbf{c}_{p}$.

\paragraph{Summary.} 
% \zhuowan{If there are important things that are missed out, put them here, \eg the method specific to scene graph perturbation (if any).}
For each context object, we get $k$ swaps for its class labels and another $k$ swaps for attributes. Thus $2k$ context swaps are performed for each irrelevant object. To generate one perturbation for an image, we only perturb one context object at a time.  Given $m$ context objects in the image, we perform $m*2k$ perturbations. This is detailed testing on the model to check if it depends on context for predictions. The results of these $m*2k$ perturbations are used to measure the robustness of the model.

\subsection{SwapMix as a training strategy.}
% \zhuowan{This subsection should be elaborated better. Put in more details, \eg first paragraph for motivation of SwapMix; second paragraph for how to identify the context and how to swap them with probability; third paragraph say we do it in scene graph and in feature spaces, in terms of object names or features; then put more details like training strategy, training time (training time will be doubled?)}
% Using SwapMix we find the robustness of the VQA models towards context swapping.
We can further use SwapMix to improve the robustness of the model. We use SwapMix during training to augment the training images. The model sees a new version of the image at every epoch based on context swapping. Using SwapMix with training, we force the model to pay less attention to context, $C$, and focus on relevant objects in the image to answer the questions.

During training, we swap the feature vectors belonging to context with other feature vectors from the dataset. We identify the context and perform context swapping based on (a) class label and (b) attributes in the same way as explained in the above sections. We perform context swapping on some irrelevant objects. For every irrelevant object, we decide to swap a feature with a probability of $p=0.5$. If selected for context swapping, we decide if we have to perform context swapping based on the class label with a probability of $p=0.5$, otherwise, we perform context swapping based on attributes.

SwapMix training can be performed on both models trained with FasterRCNN features and model with perfect sight. We add a new function in the data loading part of training, which changes the context in the image. As we do context swapping for every image during data loading, the training time increases by a factor of 1.4 times. In our analysis, we show that both the robustness of the model and the effective accuracy increases using SwapMix on both FasterRCNN and Perfect Sight embeddings.

% We filter objects in the image which are not irrelevant to the question. We then find the corresponding faster-rcnn feature to irrelevant objects using Intersection-over-Union(IoU) ratio between gqa annotation and faster rcnn output. 

% For each question, we swap the faster-rcnn feature corresponding to a irrelevant object with a new feature using on our knowledge base. 
% In our knowledge base, each object is mapping to its similar object. For example, car is mapped to truck, bus, train etc. Based on the number of similar objects in the knowledge base to the irrelevant objects, we perform 2 sets of experiments. One where we perform atmost N feature swaps for each irrelevant object, and one where we perform exactly N feature swaps for each irrelevant object.  

% In this section, we first describe our experimental setup on GQA dataset. We provide analysis on 2 types of VQA models, MCAN and LXMERT. Then we provide quantitative analysis to expose the robustness of VQA models.

%% file: content/04_experiments.tex
% \FloatBarrier 
\begin{table*}[!ht]
\begin{center}
\vspace{-0.5em}
\begin{tabular}{l|ccc|ccc}
\toprule
                      & \multicolumn{3}{c|}{\bf{MCAN}}                           & \multicolumn{3}{c}{\bf{LXMERT}}             \\
                      &     Acc.   & Context Reliance   &     Effective Acc.           &     Acc.     & Context Reliance    & Effective Acc.  \\ \hline
Faster RCNN            &    70.55  &           45.05       &      38.77            &   83.78 \protect\footnotemark    &      10.10             &       75.32    \\
Perfect Sight          &    90.34  &       16.40           &       75.53           &   91.58     &      18.85             &       74.31        \\ \hline
Faster RCNN + SwapMix    &    61.04  &          26.94        &       44.61           &   83.72     &      7.31             &       77.60         \\ 
Perfect Sight + SwapMix  &    88.10  &     11.65             &        77.83          &   91.45     &      17.34             &       75.59           \\ \bottomrule
\end{tabular}
\end{center}
\vspace{-1.0em}
\caption{Results for diagnosing the context reliance for MCAN \cite{yu2019deep} and LXMERT \cite{tan2019lxmert} models. We study models trained with both FasterRCNN features and perfect sight embeddings. Here \textit{Context Reliance} is the percentage of correctly-answered questions that are successfully perturbed by SwapMix; \textit{Effective Acc.} is the context-robust accuracy.}
\label{tab:overall_results}
\vspace{-1.0em}
\end{table*}
% \FloatBarrier 

\footnotetext{We note that LXMERT finetuned with Faster RCNN features has higher accuracy and shows high robustness towards SwapMix perturbation. This is because we test both the models on GQA val split and LXMERT was pretrained with five large vision-language datasets where it has seen images in GQA val set during pretraining. This is reported in the codebase. Therefore the LXMERT results with Faster RCNN features needs to be viewed with cautious.}

\section{Experiment}

\subsection{Dataset and Experiment Setup}
% \vipul{Para1: 1-2 lines on GQA dataset and how annotations are useful for feature swapping. We test on finetune and swapmix techniques on 2 VQA models - MCAN and LxMERT.}
\paragraph{Dataset.}
Our experiments are based on the GQA dataset \cite{GQA}. GQA train split contains 72140 images with 943k questions and val split contains 10243 images with 132k questions. The dataset provides annotated scene graphs for each image and ground-truth reasoning steps for each question. We leverage the reasoning steps to identify visual context and leverage the scene graph annotation to train models with perfect sight. We train the models on GQA train set and test them on GQA val test. GQA also has a test-dev split and a test split, which are not used in our work because they do not have scene graph and reasoning step annotation.

% \vipul{Para2: Have used the default learning rate and other parameters as describe by authors in the paper. Learning rate = ... for MCAN and .. for LxMERT}
\paragraph{Models.}
Among the many different VQA models, in this work, we focus on two representative models: MCAN \cite{yu2019deep} and LXMERT \cite{tan2019lxmert}. MCAN is a representative of attention-based models, which contains self-attention and guided-attention units to model the intra-modal and inter-modal interactions between visual and question input. LXMERT is a representative of large-scale pretrained models which can be then finetuned to solve a set of downstream tasks.

\paragraph{Implementation Details.}
We use the official released code for both MCAN and LXMERT models. We finetune both MCAN and LXMERT pre-trained models using FasterRCNN features on the GQA train set using the default hyperparameters as described by respective authors. For training models with perfect sight, we get ground-truth object names and attributes from scene graph annotation in GQA dataset. MCAN with perfect sight takes a total of 50 epochs to converge and LXMERT model takes 6 epochs. For LXMERT, we use the object features provided by its authors in the official codebase. For MCAN, we used the object features released with GQA dataset. %\zhuowan{Put links for MCAN repo, LXMERT repo and repo for FasterRCNN features.}

\paragraph{Evaluation metrics.}
We introduce 2 new metrics to evaluate the model robustness, \textit{context reliance} and \textit{effective accuracy}. As explained in Section \ref{sec:swapmix}, we apply $2mk$ perturbations for each question where $m$ is the number of irrelevant objects in the image and $2k$ is the number of feature swaps per irrelevant object. We consider a question relying on context if its answer changes for any for the $2mk$ perturbations. Based on this definition, \textit{context reliance} is the percentage of context-relying questions that are originally answered correctly. \textit{Effective Accuracy} is the percentage of questions that are consistently predicted correctly and survive all $2mk$ SwapMix perturbations. Mathematically, it can be written as \textit{effective Acc} = ${\sum^{N}_{i} q_{i}}/{N}$, where N is the total number of questions in the dataset and $q_{i}$ is defined as:
\begin{align*}
     q_{i} = \begin{dcases*}
        0, & if $ gt  \neq Model(V^{j},Q)$ for any perturbation j  \\
        1, & otherwise. 
        \end{dcases*}
\end{align*}

% For all our experiments,  we report results only for the correctly answered questions without perturbations.

\begin{table*}[ht]
\begin{center}
\begin{tabular}{l|l|cc|cc}
\toprule
                         &                      &         \multicolumn{2}{c|}{\bf{MCAN}}        & \multicolumn{2}{c}{\bf{LXMERT}}            \\
                         &                      &  Class Reliance  & Attr Reliance &  Class Reliance & Attr Reliance \\ \hline
\multirow{4}{*}{k=5}     &  Faster RCNN          &       32.52       &    28.41         &       7.29        &   7.61         \\
                         &  Faster RCNN + SwapMix  &       18.19       &    17.10         &       5.78        &   5.94         \\ \cline{2-6} 
                         &  Perfect Sight       &       11.11       &    3.16          &       14.57       &    1.34           \\ 
                         &  Perfect Sight + SwapMix&       7.64        &    2.36          &       12.89       &    1.32           \\ \hline
                         
\multirow{4}{*}{k=10}    &  Faster RCNN          &       39.40       &    34.47         &       8.18        &   8.81          \\
                         &  Faster RCNN + SwapMix  &       22.18       &    20.91         &       6.27        &   6.58        \\ \cline{2-6}
                         &  Perfect Sight        &       15.52       &    3.71          &       18.82       &   1.35            \\ 
                         &  Perfect Sight + SwapMix&       10.89       &    2.72          &       17.28       &   1.36            \\ \bottomrule
\end{tabular}
\end{center}
\vspace{-1.0em}
\caption{Detailed analysis of reliance on context. Here we measure the reliance on (a) class names and (b) attributes of irrelevant objects on model prediction. We provide analysis on $k$ perturbations on context for each irrelevant object.}
\vspace{-1.0em}
\label{tab:detailed_results}

\end{table*}

\subsection{SwapMix Perturbation Results}
% \zhuowan{I see the results of LXMERT. The \% of changes is much lower than LXMERT. Does it mean LXMERT is more robust than MCAN?} \vipul{Yes, these numbers mean that LXMERT is more robust. Which is consistent with the fact that big transformer models are more robust than attention models.}

We finetune the MCAN model and LXMERT model on GQA training split with object features extracted by pretrained Faster RCNN. After finetuning, MCAN reaches 70.55\% accuracy and LXMERT reaches 83.78\% accuracy on GQA validation split. These results are comparable with ones reported by original authors. 
% \zhuowan{Is MCAN finetuned from VQAv2 pretrained model or trained from scratch?} \vipul{It is finetuned from VQAv2 pretrained model}.
% \zhuowan{@vipul, any justification to show that the accuracy is comparable with ones reported in the original papers?} \vipul{Because we have used the same hyperparameters as set by the authors. I checked the LXMERT and MCAN numbers from various github comments and they were comparable}

Then we perform SwapMix perturbation to extensively test models' reliance on context. The evaluation results for both the MCAN model and LXMERT model are shown in Table \ref{tab:overall_results}. For measuring robustness, context reliance and the effective accuracy are reported. 
% Context reliance is the percentage of correctly-answered questions that get wrong after perturbation, while effective accuracy is the question answering accuracy that is robust to SwapMix perturbation. 
Surprisingly, 45\% of MCAN answers get changed after perturbation and the effective accuracy drops significantly from 70.55\% to 38.77\%. The significant drop suggests that the MCAN model relies heavily on the context and is not robust to context swapping. On the contrary, LXMERT is more robust. We conjecture this is because LXMERT is pretrained on a large amount of image-text pairs from five vision-and-language datasets \cite{tan2019lxmert} and the large-scale pretraining equipped the model with better robustness.

% on object-based features extracted by object detector FasterRCNN and scene graph embeddings for perfect sight. The accuracy of both the models were 70.55\% and 83.78\% after finetuning. 

% Table \ref{tab:overall_results} shows results for SwapMix on MCAN and LXMERT models on FasterRCNN and scene graph embeddings. The column Percent reliant on context represents the percentage of question which were affected by SwapMix perturbations. These questions were affected either by (a)class name or (b)attribute perturbation. We test the models on GQA val set.  We use LXMERT to test the correctness of using SwapMix as a data augmentation tool and models trained with perfect sight.

Next, we study VQA models with perfect sight. We train both models with perfect sight using encodings of ground truth object names and attributes. As shown in table \ref{tab:overall_results}, both models achieve more than 90\% accuracy with perfect sight. We observe a significant improvement in robustness of MCAN with perfect sight: its context reliance drops by 28.7\% compared with training on Faster RCNN features (from 45.1\% to 16.4\%) and the effective accuracy improves from 38.77\% to 75.53\%. This suggests that models trained with perfect sight are more robust than its FasterRCNN counterparts when trained with the same amount of data. It is also noticeable that the LXMERT performance is in a similar range with MCAN, which suggests that LXMERT is no more robust than MCAN without seeing more pretraining data in the same domain.

In Table \ref{tab:detailed_results}, we provide more detailed results of perturbations on object class and attribute separately. Interestingly, we observe that models with perfect sight are highly robust to \textit{attribute} perturbations: only 3.7\% and 1.4\% of the correct answers get changed by attribute perturbation for MCAN and LXMERT respectively. This suggests that given the ground-truth class name, the model can distinguish the relevant and irrelevant objects well, thus are robust to perturbation on the attributes of context objects. 

To further support our claim of generalisation of SwapMix, we tested our approach on OSCAR \cite{li2020oscar}. We see 26.3\% of OSCAR answers relies on visual context. The results are consistent with our initial results that transformer models are more robust than MCAN.

\subsection{SwapMix Training Results}

Using SwapMix as a training data augmentation strategy consistently improves the robustness of both models in all settings. For both MCAN and LXMERT, trained with both FasterRCNN features and perfect encodings, SwapMix training reduces the models' reliance on context and boosts the effective accuracy. 

As shown in Table \ref{tab:overall_results} (marked as +SwapMix), SwapMix training significantly decreases the context reliance of MCAN by 40\% (from 45\% to 27\%) and increases its effective accuracy by 5.8\% (row 3). The results are also consistent for MCAN with perfect sight and LXMERT. Table \ref{tab:detailed_results} further shows that SwapMix training improves robustness in both context class reliance and attribute reliance. The results consistently show that SwapMix as a training strategy decreases model reliance on context, encourages model robustness and improves effective accuracy. 

Interestingly, we notice that there is a trade-off between model robustness and overall accuracy. While we see significant improvement in model robustness, it is noticeable that the overall model accuracy drops to some extent. For example, when applying SwapMix training to MCAN model with perfect sight, its context reliance reduces by 4.7\% and effective accuracy improves by 2.3\%, while the overall model accuracy drops by 2.2\%. The model utilizes biases and correlations in context to achieve high performance, thus when the context reliance is reduced by SwapMix training, the effective accuracy improves while the overall accuracy drops. Hereby we suggest that the effective accuracy is a better description of the models' true ability to understand the task without relying on context bias.

\subsection{Ablations and Analysis}
\paragraph{Ablating the swapping number $k$.} In Table \ref{tab:detailed_results}, we additionally provide ablation study results for the hyperparameter $k$, which is the perturbation number. The results for $k=5$ or $k=10$ are shown in the table. We do $k$ perturbations on class names and attributes of context objects and report the percentage of questions affected. The result shows that when we increase the perturbations number of $k$ from 5 to 10, the reported answer changes increase accordingly for both models, which is expected.  Whereas it is also notable that the reliance increase is not significant, showing that most reliance on context can be revealed with a relatively small number of perturbations. By default, we use $k$=10 to benchmark reliance on the context of VQA models. 

% Training models with SwapMix data augmentation improves the robustness for both class name and attribute perturbations on irrelevant objects.
%  SwapMix helps the model focus less of context and thus making it more robust.

\paragraph{Random padding to $k$ swaps.} When doing object name perturbation, for cases where number of compatible classes is less than $k$, we select random classes from the dataset to pad the perturbation number to exactly $k$. To verify that this random padding does not bring extra noise in the result, we compare results with and without random padding. As shown in table \ref{tab:random_show}, 
% the context reliance changes consistently with and without random padding, which verifies that 
the effect of random padding is negligible.

\begin{table}[ht]
\centering
\begin{tabular}{l|l|c|c}
\toprule 
            \multicolumn{4}{c}{MCAN}     \\ \hline
                        &                  &          Random          &  w/o random   \\ \hline
\multirow{2}{*}{k=10}   &  FRCNN           &           45.1           &       40.3     \\
                        &  FRCNN + SwapMix &           26.9           &       24.3     \\ \hline
\multirow{2}{*}{k=5}    &  FRCNN           &           38.1           &       35.0     \\
                        &  FRCNN + SwapMix &           22.4           &       20.8     \\ \bottomrule
\end{tabular}
\caption{Context reliance measured by SwapMix with and without random padding to generate $k$ perturbations. This table verifies that random padding does not lead to significant difference.}
\vspace{-1.5em}
\label{tab:random_show}
\end{table}

\begin{figure}[b]
\vspace{-0.5em}
\begin{center}
% \fbox{\rule{0pt}{2in} \rule{.9\linewidth}{0pt}}
\includegraphics[width=1.0\linewidth]{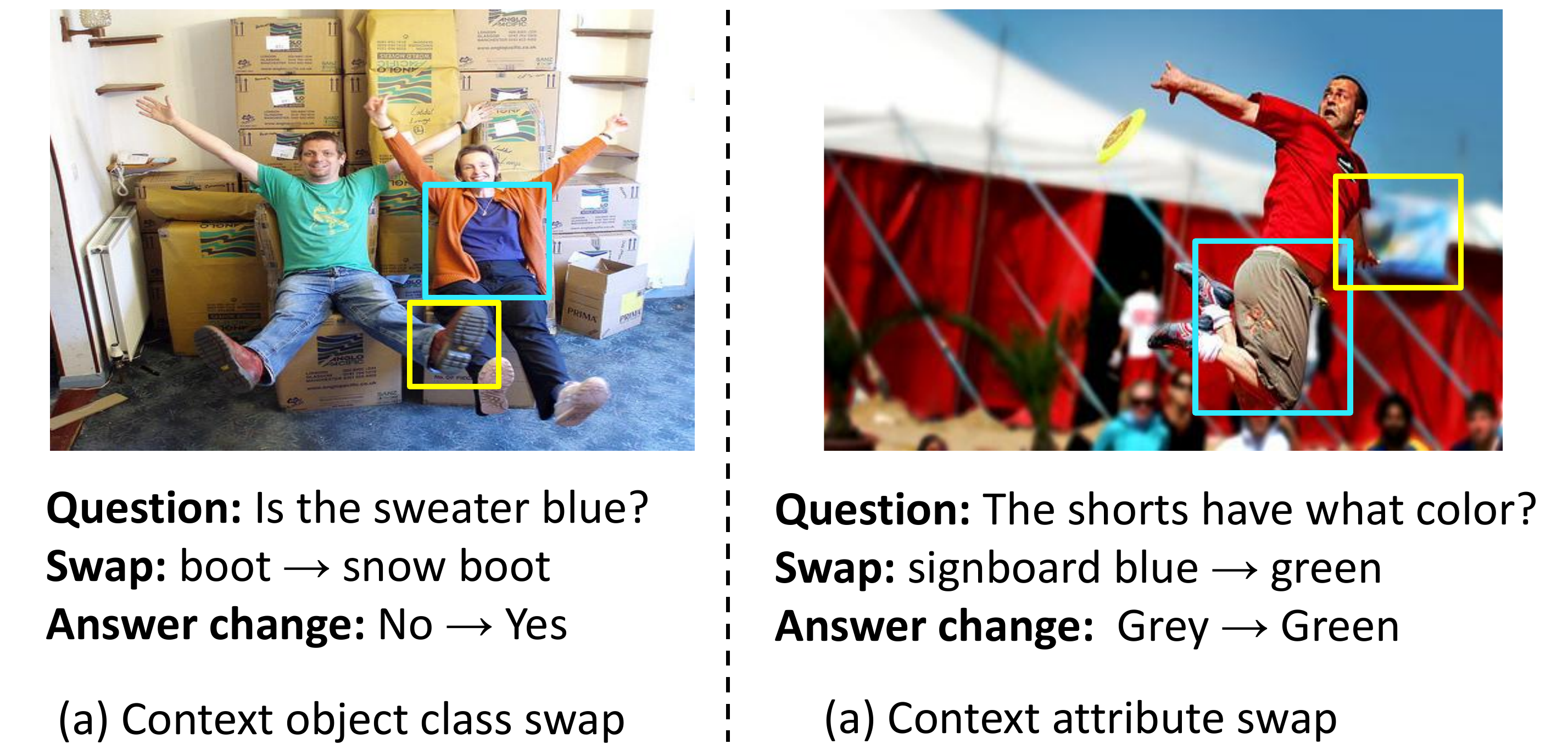}
\end{center}
\vspace{-1.5em}
  \caption{Examples for SwapMix perturbationson GQA val split. \textcolor{cyan}{Blue} boxes show relevant objects and \textcolor{yellow}{yellow} boxes show context objects. In (a) we perform class name perturbation and change boots to snow boots. In (b) we perform attribute perturbation and change color of signboard from blue to green. Both these SwapMix perturbations change model prediction. }
\vspace{-0.5em}
\label{fig:examples}
\end{figure}

\paragraph{Examples for SwapMix Perturbation.}
In Figure \ref{fig:examples}, we show examples for our proposed SwapMix perturbation. Example (a) is based on class name swapping and example (b) is based on attribute swapping, both of which resulted in the change of model prediction. In example (a), the boot is irrelevant to the question about sweater color, while changing boots into snow boots results in a change in model prediction. In example (b), when we swap the blue signboard in the background with a green signboard, the model prediction on the short's color changed to green as well. The examples are based on the results of MCAN model with perfect sight. The examples intuitively show that VQA models rely heavily on context and by perturbing irrelevant context in the image, we can change model prediction. 
% Objects which are relevant to the question are marked by blue bounding box and the irrelevant object used for feature swapping is marked by yellow bounding box. This shows that, VQA models rely on context for predictions. We analyze the impact of SwapMix on GQA validation images. We are able to change predictions using (a) class name and (b) attribute perturbation on irrelevant objects. 

\begin{figure}[t]
\vspace{-0.5em}
\begin{center}
% \fbox{\rule{0pt}{2in} \rule{.9\linewidth}{0pt}}
% \includegraphics[width=1.0\linewidth]{figures/Motivation.pdf}
\includegraphics[width=1.0\linewidth]{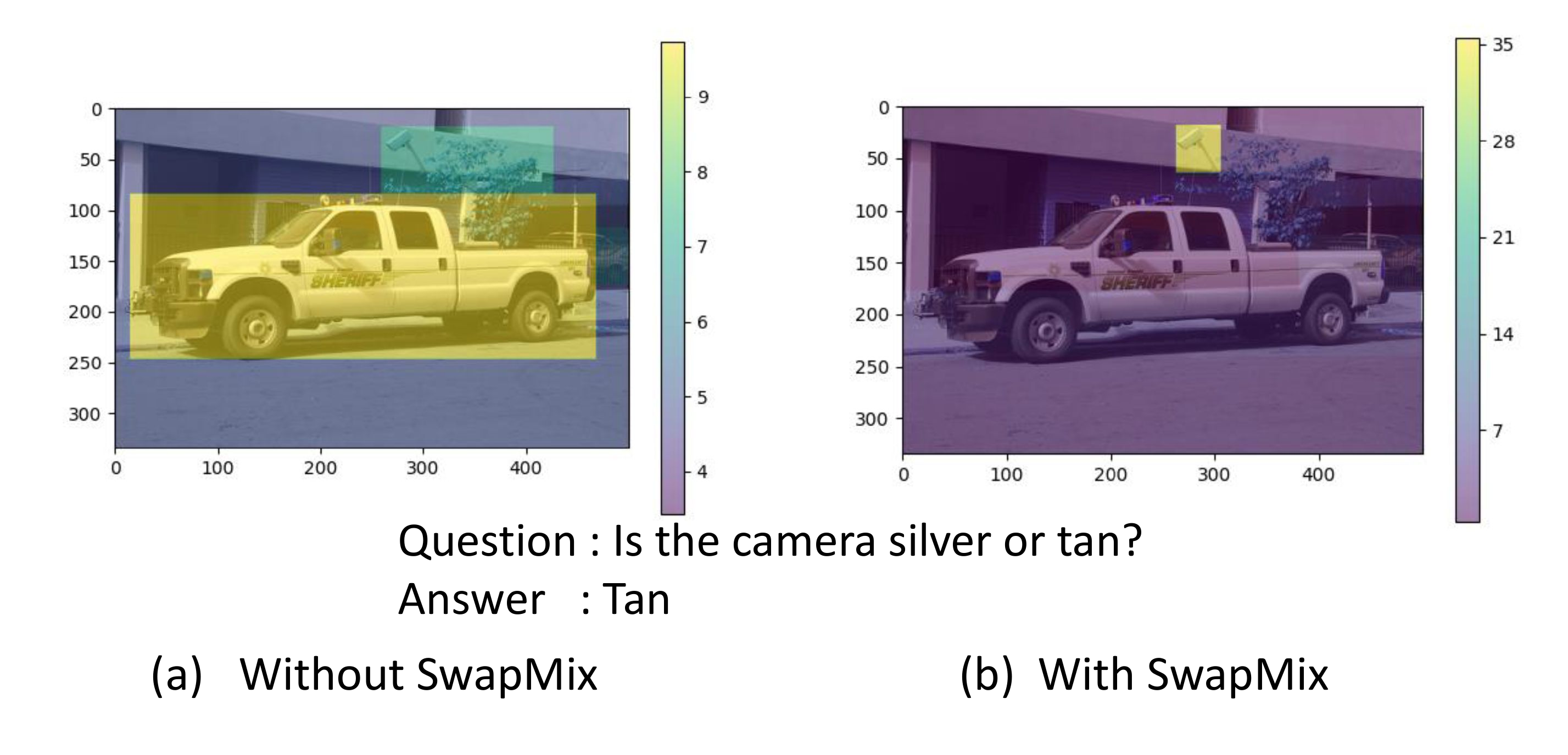}\end{center}
\vspace{-1.5em}
  \caption{Visualization of attention weights for models (a) without SwapMix training (b) with SwapMix training. SwapMix training effectively suppresses models' reliance on visual context.}
\vspace{-1.5em}
\label{fig:attention}
\end{figure}

\paragraph{Attention visualization for SwapMix training.}
Training using SwapMix as data augmentation reduces the models' reliance on context. In Figure \ref{fig:attention}, we show the visualization of attention weights for models trained without SwapMix and models trained using SwapMix as data augmentation. The visualization is based on the LXMERT model with perfect sight. For the given question, ``Is the camera silver or tan", a model with vanilla training pays more attention to irrelevant context objects such as the car, the tree, etc., while model trained with SwapMix augmentation focuses highly on the relevant object, camera, and pays very little attention to other objects. The visualization qualitatively shows that when applied as data augmentation strategy, SwapMix effectively suppresses the model's dependency on visual context and forces the model to focus more on relevant objects. 
% By doing so, SwapMix improves robustness, reduces reliance on irrelevant context and improves effective accuracy for all types of VQA modals.  
% This proves that SwapMix helps the model to focus more on relevant objects in the image.

%% file: content/05_conclusion.tex
\section{Conclusion}
In this work, we study the reliance of VQA models on context, i.e. irrelevant objects in the image for prediction. We propose a simple yet effective perturbation technique: SwapMix. SwapMix is effective in both diagnosing model robustness on context reliance, and regularizing the context reliance of VQA models thus making them more robust. Our experiments of two representative models on GQA show the effectiveness of SwapMix. Interestingly, we find that the robustness of VQA models highly depends on the quality of visual perception and models with perfect sight are more robust to context perturbation. Large-scale pretraining also helps improve model robustness. We hope that our initial analysis on reliance on visual context can serve as a starting point for future researchers to study VQA robustness and reliability. 

\paragraph{Negative impact and limitations.} Our work study the robustness of VQA models and find that the models are vulnerable to context perturbations. The proposed SwapMix perturbation strategy may be used maliciously to attack VQA models. To overcome this potential negative impact, we suggest that training with SwapMix can effectively regularize reliance on context and that better visual representation may improve model robustness. The limitation of our work is that we only study two representative models, using two types of visual features on the GQA dataset. 
% More experiments with more extensive coverage of models and visual representations will strengthen the findings of our work.

%% file: content/06_acknowledgments.tex
\section*{Acknowledgements}
This work is supported by NSF 1763705.

%% file: main.bbl
\begin{thebibliography}{10}\itemsep=-1pt

\bibitem{agrawal2016analyzing}
Aishwarya Agrawal, Dhruv Batra, and Devi Parikh.
\newblock Analyzing the behavior of visual question answering models.
\newblock {\em arXiv preprint arXiv:1606.07356}, 2016.

\bibitem{VQA-CP_v2}
Aishwarya Agrawal, Dhruv Batra, Devi Parikh, and Aniruddha Kembhavi.
\newblock Don't just assume; look and answer: Overcoming priors for visual
  question answering.
\newblock In {\em Proceedings of the IEEE Conference on Computer Vision and
  Pattern Recognition}, pages 4971--4980, 2018.

\bibitem{anderson2018bottom}
Peter Anderson, Xiaodong He, Chris Buehler, Damien Teney, Mark Johnson, Stephen
  Gould, and Lei Zhang.
\newblock Bottom-up and top-down attention for image captioning and visual
  question answering.
\newblock In {\em Proceedings of the IEEE conference on computer vision and
  pattern recognition}, pages 6077--6086, 2018.

\bibitem{antol2015vqa}
Stanislaw Antol, Aishwarya Agrawal, Jiasen Lu, Margaret Mitchell, Dhruv Batra,
  C~Lawrence Zitnick, and Devi Parikh.
\newblock Vqa: Visual question answering.
\newblock In {\em Proceedings of the IEEE international conference on computer
  vision}, pages 2425--2433, 2015.

\bibitem{bahdanau2019closure}
Dzmitry Bahdanau, Harm de Vries, Timothy~J O'Donnell, Shikhar Murty, Philippe
  Beaudoin, Yoshua Bengio, and Aaron Courville.
\newblock Closure: Assessing systematic generalization of clevr models.
\newblock {\em arXiv preprint arXiv:1912.05783}, 2019.

\bibitem{ben2017mutan}
Hedi Ben-Younes, R{\'e}mi Cadene, Matthieu Cord, and Nicolas Thome.
\newblock Mutan: Multimodal tucker fusion for visual question answering.
\newblock In {\em Proceedings of the IEEE international conference on computer
  vision}, pages 2612--2620, 2017.

\bibitem{Cadne2019MURELMR}
R{\'e}mi Cad{\`e}ne, Hedi Ben-younes, Matthieu Cord, and Nicolas Thome.
\newblock Murel: Multimodal relational reasoning for visual question answering.
\newblock {\em 2019 IEEE/CVF Conference on Computer Vision and Pattern
  Recognition (CVPR)}, pages 1989--1998, 2019.

\bibitem{RUBi}
Remi Cadene, Corentin Dancette, Matthieu Cord, Devi Parikh, et~al.
\newblock Rubi: Reducing unimodal biases for visual question answering.
\newblock {\em Advances in neural information processing systems}, 32:841--852,
  2019.

\bibitem{cascante2022simvqa}
Paola Cascante-Bonilla, Hui Wu, Letao Wang, Rogerio Feris, and Vicente Ordonez.
\newblock Simvqa: Exploring simulated environments for visual question
  answering.
\newblock In {\em Proceedings of the IEEE Conference on Computer Vision and
  Pattern Recognition (CVPR)}, 2022.

\bibitem{chen2015abc}
Kan Chen, Jiang Wang, Liang-Chieh Chen, Haoyuan Gao, Wei Xu, and Ram Nevatia.
\newblock Abc-cnn: An attention based convolutional neural network for visual
  question answering.
\newblock {\em arXiv preprint arXiv:1511.05960}, 2015.

\bibitem{CSS}
Long Chen, Xin Yan, Jun Xiao, Hanwang Zhang, Shiliang Pu, and Yueting Zhuang.
\newblock Counterfactual samples synthesizing for robust visual question
  answering.
\newblock In {\em Proceedings of the IEEE/CVF Conference on Computer Vision and
  Pattern Recognition}, pages 10800--10809, 2020.

\bibitem{clark2019don}
Christopher Clark, Mark Yatskar, and Luke Zettlemoyer.
\newblock Don't take the easy way out: Ensemble based methods for avoiding
  known dataset biases.
\newblock {\em arXiv preprint arXiv:1909.03683}, 2019.

\bibitem{Dancette2021BeyondQB}
Corentin Dancette, R{\'e}mi Cad{\`e}ne, Damien Teney, and Matthieu Cord.
\newblock Beyond question-based biases: Assessing multimodal shortcut learning
  in visual question answering.
\newblock {\em 2021 IEEE/CVF International Conference on Computer Vision
  (ICCV)}, pages 1554--1563, 2021.

\bibitem{fukui2016multimodal}
Akira Fukui, Dong~Huk Park, Daylen Yang, Anna Rohrbach, Trevor Darrell, and
  Marcus Rohrbach.
\newblock Multimodal compact bilinear pooling for visual question answering and
  visual grounding.
\newblock {\em arXiv preprint arXiv:1606.01847}, 2016.

\bibitem{gokhale2020mutant}
Tejas Gokhale, Pratyay Banerjee, Chitta Baral, and Yezhou Yang.
\newblock Mutant: A training paradigm for out-of-distribution generalization in
  visual question answering.
\newblock {\em arXiv preprint arXiv:2009.08566}, 2020.

\bibitem{goyal2017making}
Yash Goyal, Tejas Khot, Douglas Summers-Stay, Dhruv Batra, and Devi Parikh.
\newblock Making the v in vqa matter: Elevating the role of image understanding
  in visual question answering.
\newblock In {\em Proceedings of the IEEE Conference on Computer Vision and
  Pattern Recognition}, pages 6904--6913, 2017.

\bibitem{he2019adaptive}
Junjun He, Zhongying Deng, Lei Zhou, Yali Wang, and Yu Qiao.
\newblock Adaptive pyramid context network for semantic segmentation.
\newblock In {\em Proceedings of the IEEE/CVF Conference on Computer Vision and
  Pattern Recognition}, pages 7519--7528, 2019.

\bibitem{hendricks2018women}
Lisa~Anne Hendricks, Kaylee Burns, Kate Saenko, Trevor Darrell, and Anna
  Rohrbach.
\newblock Women also snowboard: Overcoming bias in captioning models.
\newblock In {\em Proceedings of the European Conference on Computer Vision
  (ECCV)}, pages 771--787, 2018.

\bibitem{GQA}
Drew~A Hudson and Christopher~D Manning.
\newblock Gqa: A new dataset for real-world visual reasoning and compositional
  question answering.
\newblock In {\em Proceedings of the IEEE/CVF conference on computer vision and
  pattern recognition}, pages 6700--6709, 2019.

\bibitem{GQA-OOD}
Corentin Kervadec, Grigory Antipov, Moez Baccouche, and Christian Wolf.
\newblock Roses are red, violets are blue... but should vqa expect them to?
\newblock In {\em Proceedings of the IEEE/CVF Conference on Computer Vision and
  Pattern Recognition}, pages 2776--2785, 2021.

\bibitem{Kervadec2021HowTA}
Corentin Kervadec, Theo Jaunet, Grigory Antipov, Moez Baccouche, Romain
  Vuillemot, and Christian Wolf.
\newblock How transferable are reasoning patterns in vqa?
\newblock {\em 2021 IEEE/CVF Conference on Computer Vision and Pattern
  Recognition (CVPR)}, pages 4205--4214, 2021.

\bibitem{kim2018bilinear}
Jin-Hwa Kim, Jaehyun Jun, and Byoung-Tak Zhang.
\newblock Bilinear attention networks.
\newblock {\em arXiv preprint arXiv:1805.07932}, 2018.

\bibitem{kim2016hadamard}
Jin-Hwa Kim, Kyoung-Woon On, Woosang Lim, Jeonghee Kim, Jung-Woo Ha, and
  Byoung-Tak Zhang.
\newblock Hadamard product for low-rank bilinear pooling.
\newblock {\em arXiv preprint arXiv:1610.04325}, 2016.

\bibitem{kortylewski2020compositional}
Adam Kortylewski, Ju He, Qing Liu, and Alan~L Yuille.
\newblock Compositional convolutional neural networks: A deep architecture with
  innate robustness to partial occlusion.
\newblock In {\em Proceedings of the IEEE/CVF Conference on Computer Vision and
  Pattern Recognition}, pages 8940--8949, 2020.

\bibitem{kortylewski2021compositional}
Adam Kortylewski, Qing Liu, Angtian Wang, Yihong Sun, and Alan Yuille.
\newblock Compositional convolutional neural networks: A robust and
  interpretable model for object recognition under occlusion.
\newblock {\em International Journal of Computer Vision}, 129(3):736--760,
  2021.

\bibitem{Li2019VisualBERTAS}
Liunian~Harold Li, Mark Yatskar, Da Yin, Cho-Jui Hsieh, and Kai-Wei Chang.
\newblock Visualbert: A simple and performant baseline for vision and language.
\newblock {\em ArXiv}, abs/1908.03557, 2019.

\bibitem{li2020oscar}
Xiujun Li, Xi Yin, Chunyuan Li, Pengchuan Zhang, Xiaowei Hu, Lei Zhang, Lijuan
  Wang, Houdong Hu, Li Dong, Furu Wei, et~al.
\newblock Oscar: Object-semantics aligned pre-training for vision-language
  tasks.
\newblock In {\em European Conference on Computer Vision}, pages 121--137.
  Springer, 2020.

\bibitem{lin2018scn}
Di Lin, Ruimao Zhang, Yuanfeng Ji, Ping Li, and Hui Huang.
\newblock Scn: switchable context network for semantic segmentation of rgb-d
  images.
\newblock {\em IEEE transactions on cybernetics}, 50(3):1120--1131, 2018.

\bibitem{lin2020gps}
Xin Lin, Changxing Ding, Jinquan Zeng, and Dacheng Tao.
\newblock Gps-net: Graph property sensing network for scene graph generation.
\newblock In {\em Proceedings of the IEEE/CVF Conference on Computer Vision and
  Pattern Recognition}, pages 3746--3753, 2020.

\bibitem{malinowski2014multi}
Mateusz Malinowski and Mario Fritz.
\newblock A multi-world approach to question answering about real-world scenes
  based on uncertain input.
\newblock {\em Advances in neural information processing systems},
  27:1682--1690, 2014.

\bibitem{manjunatha2019explicit}
Varun Manjunatha, Nirat Saini, and Larry~S Davis.
\newblock Explicit bias discovery in visual question answering models.
\newblock In {\em Proceedings of the IEEE/CVF Conference on Computer Vision and
  Pattern Recognition}, pages 9562--9571, 2019.

\bibitem{million2007hadamard}
Elizabeth Million.
\newblock The hadamard product.
\newblock 2007.

\bibitem{Niu2021CounterfactualVA}
Yulei Niu, Kaihua Tang, Hanwang Zhang, Zhiwu Lu, Xiansheng Hua, and Ji-Rong
  Wen.
\newblock Counterfactual vqa: A cause-effect look at language bias.
\newblock {\em 2021 IEEE/CVF Conference on Computer Vision and Pattern
  Recognition (CVPR)}, pages 12695--12705, 2021.

\bibitem{Niu_2021_CVPR}
Yulei Niu, Kaihua Tang, Hanwang Zhang, Zhiwu Lu, Xian-Sheng Hua, and Ji-Rong
  Wen.
\newblock Counterfactual vqa: A cause-effect look at language bias.
\newblock In {\em Proceedings of the IEEE/CVF Conference on Computer Vision and
  Pattern Recognition (CVPR)}, pages 12700--12710, June 2021.

\bibitem{oliva2007role}
Aude Oliva and Antonio Torralba.
\newblock The role of context in object recognition.
\newblock {\em Trends in cognitive sciences}, 11(12):520--527, 2007.

\bibitem{pennington2014glove}
Jeffrey Pennington, Richard Socher, and Christopher~D Manning.
\newblock Glove: Global vectors for word representation.
\newblock In {\em Proceedings of the 2014 conference on empirical methods in
  natural language processing (EMNLP)}, pages 1532--1543, 2014.

\bibitem{ramakrishnan2018overcoming}
Sainandan Ramakrishnan, Aishwarya Agrawal, and Stefan Lee.
\newblock Overcoming language priors in visual question answering with
  adversarial regularization.
\newblock {\em arXiv preprint arXiv:1810.03649}, 2018.

\bibitem{ren2020scene}
Guanghui Ren, Lejian Ren, Yue Liao, Si Liu, Bo Li, Jizhong Han, and Shuicheng
  Yan.
\newblock Scene graph generation with hierarchical context.
\newblock {\em IEEE Transactions on Neural Networks and Learning Systems},
  32(2):909--915, 2020.

\bibitem{Ren2015FasterRT}
Shaoqing Ren, Kaiming He, Ross~B. Girshick, and Jian Sun.
\newblock Faster r-cnn: Towards real-time object detection with region proposal
  networks.
\newblock {\em IEEE Transactions on Pattern Analysis and Machine Intelligence},
  39:1137--1149, 2015.

\bibitem{selvaraju2019taking}
Ramprasaath~R Selvaraju, Stefan Lee, Yilin Shen, Hongxia Jin, Shalini Ghosh,
  Larry Heck, Dhruv Batra, and Devi Parikh.
\newblock Taking a hint: Leveraging explanations to make vision and language
  models more grounded.
\newblock In {\em Proceedings of the IEEE/CVF International Conference on
  Computer Vision}, pages 2591--2600, 2019.

\bibitem{sharma2021visual}
Himanshu Sharma and Anand~Singh Jalal.
\newblock Visual question answering model based on graph neural network and
  contextual attention.
\newblock {\em Image and Vision Computing}, 110:104165, 2021.

\bibitem{tan2019lxmert}
Hao Tan and Mohit Bansal.
\newblock Lxmert: Learning cross-modality encoder representations from
  transformers.
\newblock {\em arXiv preprint arXiv:1908.07490}, 2019.

\bibitem{torralba2003context}
Antonio Torralba, Kevin~P Murphy, William~T Freeman, and Mark~A Rubin.
\newblock Context-based vision system for place and object recognition.
\newblock In {\em Computer Vision, IEEE International Conference on}, volume~2,
  pages 273--273. IEEE Computer Society, 2003.

\bibitem{vaswani2017attention}
Ashish Vaswani, Noam Shazeer, Niki Parmar, Jakob Uszkoreit, Llion Jones,
  Aidan~N Gomez, {\L}ukasz Kaiser, and Illia Polosukhin.
\newblock Attention is all you need.
\newblock In {\em Advances in neural information processing systems}, pages
  5998--6008, 2017.

\bibitem{wang2020robust}
Angtian Wang, Yihong Sun, Adam Kortylewski, and Alan~L Yuille.
\newblock Robust object detection under occlusion with context-aware
  compositionalnets.
\newblock In {\em Proceedings of the IEEE/CVF Conference on Computer Vision and
  Pattern Recognition}, pages 12645--12654, 2020.

\bibitem{wang2020sketching}
Wenbin Wang, Ruiping Wang, Shiguang Shan, and Xilin Chen.
\newblock Sketching image gist: Human-mimetic hierarchical scene graph
  generation.
\newblock In {\em Computer Vision--ECCV 2020: 16th European Conference,
  Glasgow, UK, August 23--28, 2020, Proceedings, Part XIII 16}, pages 222--239.
  Springer, 2020.

\bibitem{wu2019self}
Jialin Wu and Raymond~J Mooney.
\newblock Self-critical reasoning for robust visual question answering.
\newblock {\em arXiv preprint arXiv:1905.09998}, 2019.

\bibitem{yang2016stacked}
Zichao Yang, Xiaodong He, Jianfeng Gao, Li Deng, and Alex Smola.
\newblock Stacked attention networks for image question answering.
\newblock In {\em Proceedings of the IEEE conference on computer vision and
  pattern recognition}, pages 21--29, 2016.

\bibitem{yu2019deep}
Zhou Yu, Jun Yu, Yuhao Cui, Dacheng Tao, and Qi Tian.
\newblock Deep modular co-attention networks for visual question answering.
\newblock In {\em Proceedings of the IEEE/CVF Conference on Computer Vision and
  Pattern Recognition}, pages 6281--6290, 2019.

\bibitem{yu2017multi}
Zhou Yu, Jun Yu, Jianping Fan, and Dacheng Tao.
\newblock Multi-modal factorized bilinear pooling with co-attention learning
  for visual question answering.
\newblock In {\em Proceedings of the IEEE international conference on computer
  vision}, pages 1821--1830, 2017.

\bibitem{yu2018beyond}
Zhou Yu, Jun Yu, Chenchao Xiang, Jianping Fan, and Dacheng Tao.
\newblock Beyond bilinear: Generalized multimodal factorized high-order pooling
  for visual question answering.
\newblock {\em IEEE transactions on neural networks and learning systems},
  29(12):5947--5959, 2018.

\bibitem{zhong2020cascade}
Qiaoyong Zhong, Chao Li, Yingying Zhang, Di Xie, Shicai Yang, and Shiliang Pu.
\newblock Cascade region proposal and global context for deep object detection.
\newblock {\em Neurocomputing}, 395:170--177, 2020.

\bibitem{zhou2015simple}
Bolei Zhou, Yuandong Tian, Sainbayar Sukhbaatar, Arthur Szlam, and Rob Fergus.
\newblock Simple baseline for visual question answering.
\newblock {\em arXiv preprint arXiv:1512.02167}, 2015.

\end{thebibliography}
